\documentclass[runningheads]{llncs}

\usepackage{eccv}

\usepackage{eccvabbrv}

\usepackage{graphicx}
\usepackage{booktabs}

\usepackage[accsupp]{axessibility}  

\usepackage{hyperref}

\usepackage{orcidlink}

\begin{document}

\title{ISAP-3D: Identity-Slot Aligned Part-Aware 3D Generation}

\titlerunning{Abbreviated paper title}

\author{Junlin Hao \and Haoshuai Fu \and Xibin Song \and Wei Li \and Ruigang Yang \and Xinggong Zhang \and Jinchuan Zhang}



\maketitle

\begin{figure}[tbh!]
  \centering
  \includegraphics[width=0.97\linewidth]{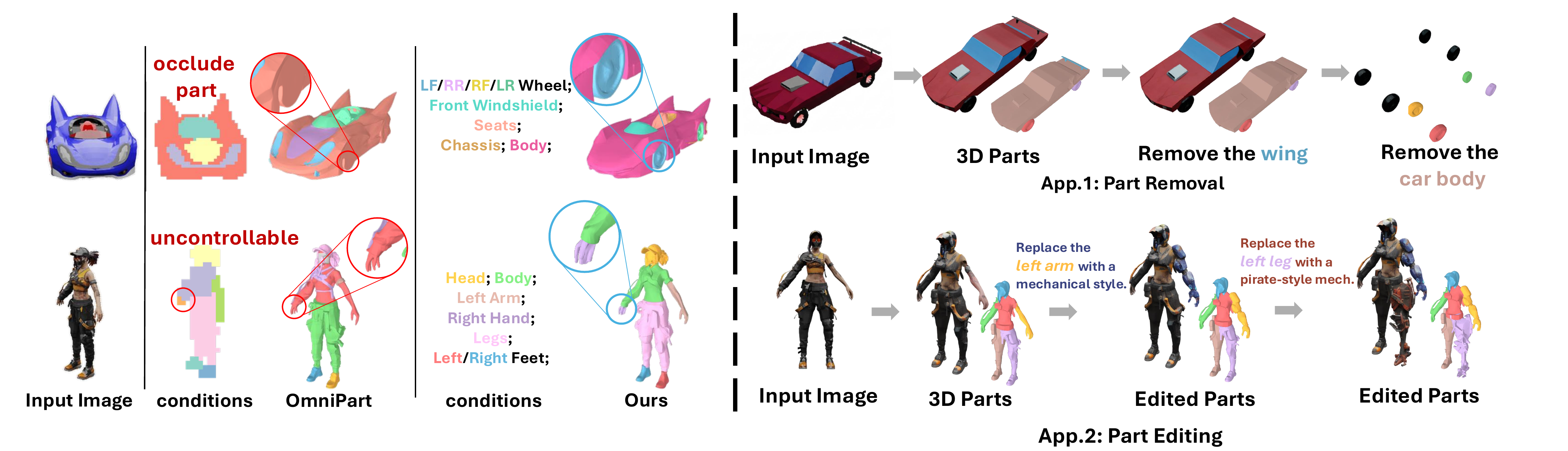}
  \caption{
  \textbf{Identity-slot aligned part-aware 3D generation.}
  Existing methods often suffer from identity–slot ambiguity, where parts may swap or merge across slots. 
  Our method explicitly aligns semantic identities with generation slots, enabling stable part synthesis from text and images and supporting intuitive part-level editing.
  }
  \label{fig:teaser}
\end{figure}

\begin{abstract}
  Part-aware 3D generation aims to synthesize structured objects with semantically meaningful components, yet often suffers from structural ambiguity due to identity-layout entanglement. 
  Existing methods either infer part identity and spatial layout implicitly, which can lead to unstable part allocation (e.g., slot swapping or part merging), or rely on strong layout conditions that are difficult to obtain in practice.
  We attribute this ambiguity to \emph{identity-slot permutation freedom}: without explicit identity-slot alignment, the correspondence between semantic parts and generation slots is not identifiable during training, allowing multiple slot assignments to fit the same supervision and leading to inconsistent decomposition.
  Based on this insight, we argue that stable part-aware generation requires \textbf{identity-aligned one-to-one slot modelling}.
  We therefore propose an identity-slot aligned framework, ISAP-3D, which anchors each part with semantic identity tokens and performs identity-conditioned one-to-one layout prediction, followed by layout-conditioned geometry synthesis.
  Structured local–global conditioning maintains identity alignment across semantic, spatial, and geometric stages.
  We also construct a part-level dataset with a unified semantic protocol to enable learnable and consistent identity-slot alignment.
  Extensive experiments demonstrate improved structural stability, controllability, and robustness over state-of-the-art part-aware generation baselines.
\end{abstract}
\section{Introduction}
\label{sec:intro}
3D generative modelling has become a key technology for scalable 3D content creation in applications such as immersive gaming, animation, and embodied AI. 
While recent advances have significantly improved holistic 3D generation quality, most approaches\cite{10.1145/3592442, li2025triposg, Xiang_2025_CVPR} still produce objects as monolithic shapes without explicit part-level structure, limiting their utility in editing, compositional reuse, and interactive manipulation. 
Part-aware 3D generation\cite{lin2025partcrafter, Chen_2025_CVPR, chen2025autopartgenautogressive3dgeneration, dong2025copart, yang2025holopart} addresses this limitation by modelling objects as assemblies of semantically meaningful components.
However, existing methods often suffer from structural instability: parts may collapse, merge, or be inconsistently decomposed even under similar conditions.
Such behaviours indicate that the correspondence between part identity and spatial layout is not reliably established during generation.
We argue that this structural ambiguity fundamentally stems from identity–layout entanglement, where semantic part identities and spatial allocation are implicitly intertwined rather than explicitly aligned.

We observe that this entanglement largely arises from the lack of explicit alignment between part identities and generation slots in existing methods. 
Many approaches\cite{lin2025partcrafter, he2025unipartpartlevel3dgeneration, tang2024partpacker, li2025moca} predict part layouts in parallel without dedicated identity anchors, allowing generation slots to compete for spatial allocation.
Others impose\cite{ding2025fullpart, yang2025omnipart} a fixed ordering strategy, such as spatial sorting, to organize the generation process. 
While such designs can regularize the spatial prediction procedure, they only constrain positional ordering rather than semantic identity.

As a consequence, the mapping between semantic parts and generation slots remains underdetermined during training.
Supervision typically constrains the overall spatial coverage of parts but does not explicitly specify which semantic identity should correspond to each generation slot. 
Moreover, training data often contains diverse part compositions under similar conditions.
Therefore, multiple slot assignments can satisfy the same supervision signal.

We refer to this phenomenon as \emph{identity–slot permutation freedom}. 
Without explicit identity–slot alignment, generation slots remain ambiguous with respect to semantic identities. 
Therefore, models may fit multiple plausible part assignments under the same condition, often relying on ordering heuristics or dataset-level composition patterns, which can lead to slot swapping, part merging, or inconsistent decompositions across samples.

Based on the above analysis, we argue that stable part-aware generation requires \textbf{identity-aligned one-to-one slot modelling}. 
Instead of allowing generation slots to implicitly compete for spatial allocation, each semantic part identity should be anchored to a dedicated slot, establishing an explicit correspondence between semantic identities and generation variables.
Such identity-aligned slots remove the interchangeability of slot assignments and ensure that part identities remain consistent throughout the generation process, independent of ordering heuristics or dataset-level priors.

Under this formulation, structured object generation can be viewed as a sequence of identity-guided reasoning steps.
Given semantic part identities, the model first predicts spatial layouts explicitly conditioned on these identity anchors, and then synthesizes detailed geometry under the predicted layouts. 
This identity-layout-geometry progression progressively reduces generation ambiguity while maintaining stable identity–slot alignment across stages.

Motivated by this insight, we design an identity-aligned generation framework, ISAP-3D, which instantiates the proposed principle in practice.
Our framework grounds generation slots with semantic identity tokens that serve as identity anchors. 
These tokens establish explicit identity–slot correspondence and guide the layout reasoning process.

Given these identity anchors, a slot-based transformer predicts spatial layouts for each part in a one-to-one manner, ensuring that layout inference is conditioned on semantic identity rather than ordering heuristics.
The predicted layouts then provide spatial constraints for the subsequent geometry synthesis stage, where detailed part geometry is generated under layout conditioning.

To maintain consistent identity grounding across stages, we further introduce a structured local–global conditioning scheme that separates part-specific signals from global contextual cues. 
This design preserves identity–slot alignment while enabling the model to incorporate complementary spatial and visual information during generation.

Achieving reliable identity–slot alignment further requires consistent semantic grounding for part identities.
While visual cues such as segmentation masks can provide strong spatial cues, they often fail to encode stable semantic identity due to occlusion or incomplete observations across views. 
As a result, relying solely on visual conditions may still leave part identities ambiguous.
To provide stable identity anchors, we adopt semantic tokens derived from textual descriptions to represent part identities. 
Text-based semantics naturally encode human-interpretable part categories and remain consistent across viewpoints and object configurations, making them well suited for identity–slot alignment.
To enable learnable and consistent semantic grounding, we construct a part-level dataset following a unified semantic protocol. 
Each token corresponds to a well-defined part identity, providing stable identity supervision for training our identity-aligned generation framework.

Our contributions are summarized as follows:
\begin{itemize}
\item We identify \emph{identity–slot permutation freedom} as a key cause of structural instability in part-aware 3D generation, where semantic identities and generation slots become implicitly interchangeable during training.

\item We propose \textbf{identity-aligned one-to-one slot modeling}, a formulation that explicitly anchors semantic part identities to dedicated generation slots, enabling stable and controllable part-aware generation.

\item We develop an identity-aligned generation framework that performs identity-conditioned layout prediction followed by layout-conditioned geometry synthesis. To facilitate learning under this formulation, we construct an identity-oriented part-level dataset with a unified semantic protocol, enabling reliable experimental validation of the proposed formulation.

\end{itemize}
\section{Related Work}
\label{sec:related}
\subsection{Holistic 3D Generation}
Holistic 3D generation aims to directly synthesize an object’s geometry and appearance. 
Existing pipelines can be broadly categorized into 2D-driven lifting approaches and native 3D generative approaches.
2D-driven methods\cite{Long_2024_CVPR,shi2023MVDream,wang2023prolificdreamerhighfidelitydiversetextto3d,10203874} leverage 2D diffusion models as priors, obtaining 3D results either through Score Distillation Sampling (SDS) optimization \cite{poole2022dreamfusiontextto3dusing2d,alldieck2024scoredistillationsamplinglearned}or via multi-view generation followed by 3D reconstruction\cite{liu2024syncdreamergeneratingmultiviewconsistentimages,liu2023zero1to3}. However, due to cross-view inconsistency inherent in 2D generation and the lack of ground-truth 3D supervision, these approaches suffer from geometric artifacts and reduced structural reliability.

With the advance of large-scale 3D datasets such as Objaverse\cite{Deitke_2023_CVPR,deitke2023objaversexluniverse10m3d}, 3D-native generative models have become increasingly feasible by learning in geometry-aware latent spaces derived from 3D assets. A common paradigm is to first train an autoencoder to compress 3D shapes into compact latents and then perform diffusion-based generation in the latent space\cite{yang2025log3dultrahighresolution3dshape,wu2025direct3ds2gigascale3dgeneration,he2025triposf}. 3DShape2VecSet\cite{10.1145/3592442}, for example, represents shapes as a set of latent tokens and introduces cross-attention for efficient set encoding. Subsequent works such as Michelangelo\cite{zhao2023michelangelo}, CLAY\cite{zhang2024clay}, Dora\cite{Chen_2025_Dora}, Hunyuan3D\cite{hunyuan3d22025tencent,hunyuan3d2025hunyuan3d}, and TripoSG\cite{li2025triposg} further improve generation fidelity, scalability, and conditioning capabilities within this latent-generation framework. More recently, methods exemplified by Trellis\cite{Xiang_2025_CVPR} and LATTICE\cite{lai2025latticedemocratizehighfidelity3d,lai2025hunyuan3d25highfidelity3d} adopt coarse-to-fine two-stage pipelines with structured latents, enabling better global structure preservation followed by local refinement, and thus more fine-grained and controllable geometric synthesis.

Despite these advances, most holistic generators still synthesize objects as monolithic shapes, lacking the explicit part-level structure and semantics needed for fine-grained editing and compositional reuse.

\subsection{Part-aware 3D Generation}
\label{sec:part_aware_generation}
Part-aware 3D generation models objects as compositions of semantically meaningful parts, enabling downstream applications such as part-level editing and re-composition.

\textbf{Implicit Part Modeling.}
A large body of work~\cite{he2025unipartpartlevel3dgeneration,li2025moca,zhu2025sealion,yang2025partdiffuserpartwise3dmesh,wang2026scenetransporteroptimaltransportguidedcompositional}
performs part generation without explicit part-level conditions, where part decomposition and assignment are entirely determined by the generative model.
PartPacker~\cite{tang2024partpacker} converts variable-part objects into a fixed-length dual-latent representation via bipartite contraction and dual volume packing, enabling compatibility with existing 3D latent denoising models.
PartCrafter~\cite{lin2025partcrafter} takes an input image with a specified number of parts and simultaneously synthesizes a part decomposition by predicting all part representations in parallel.
However, since semantic identity and spatial layout are typically modeled jointly without explicit disentanglement, these methods often struggle to ensure stable part-level semantic consistency and controllability.

\textbf{Part-conditioned Generation.}
Another line of work~\cite{Zhang_2025,ding2025muses, dong2025copart} improves generation stability and controllability by injecting part-level conditions.
One direction relies on strong layout cues.
X-Part~\cite{yan2025xpart} uses part bounding boxes as coarse spatial priors and injects point-wise semantic features from P3-SAM~\cite{ma2025p3sam} to encourage boundary-aligned parts,
while HoloPart~\cite{yang2025holopart} performs diffusion-based amodal completion of incomplete part segments to recover consistent full-part geometry.
Another direction~\cite{Chen_2025_CVPR, yang2025omnipart, chen2025autopartgenautogressive3dgeneration} leverages 2D visual cues as identity signals for layout prediction.
OmniPart~\cite{yang2025omnipart} predicts part bounding boxes using an autoregressive planner guided by image masks and then synthesizes parts jointly with a layout-conditioned 3D generator.
AutoPartGen~\cite{chen2025autopartgenautogressive3dgeneration} further generates parts sequentially in a compositional latent space.
Although these approaches reduce identity–layout ambiguity through external conditions, they often rely on strong layout annotations or noisy 2D cues that may be unreliable under occlusion or incomplete observations.

Taken together, these limitations motivate us to design an \textbf{identity-slot aligned one-to-one generation framework}.
\section{Method}

\begin{figure}[tb]
  \centering
  \includegraphics[width=0.97\linewidth]{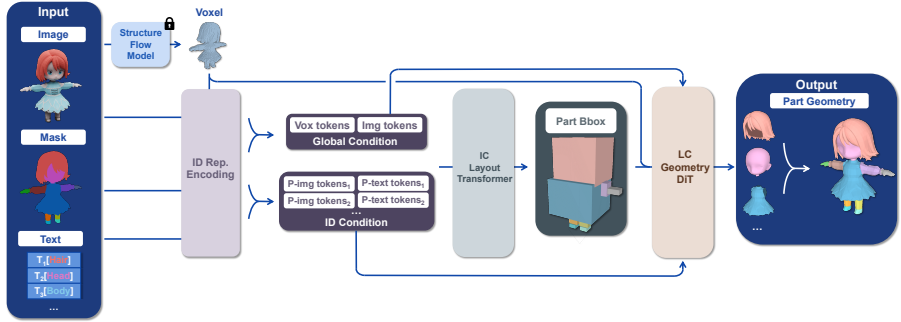}
  \caption{
   Overview of our identity-aligned part generation framework.
   Given a reference image and semantic part descriptions, we first construct identity-aware condition tokens encoding part identities and global context.
   An \textbf{I}dentity-\textbf{C}onditioned transformer predicts spatial layouts for each part, which are then used by a \textbf{L}ayout-\textbf{C}onditioned geometry synthesis network to generate part geometries.
   }
  \label{fig:overview}
\end{figure}

\subsection{Problem Formulation}
\label{subsec:formulation}
We formulate part-aware 3D generation as an identity-aligned structured generation problem, where each semantic part identity explicitly guides the generation of its corresponding 3D component.
Given a set of conditions $\mathcal{C} = \{T_k, I, M_k\}_{k=1}^{K}$, where $T_k$ denotes the semantic text description of the $k$-th part, $I$ denotes a reference image, $M_k$ denotes optional part masks, and $K$ is the number of semantic parts, the goal is to synthesize a set of per-part geometries $\mathcal{G} = \{G_k\}_{k=1}^{K}$, where each geometry $G_k$ is semantically aligned with its corresponding inputs.

To facilitate spatial reasoning, we introduce intermediate spatial representations including a coarse voxel prior $V$ inferred from the reference image $I$, and per-part bounding boxes $B = \{B_k\}_{k=1}^{K}$ that describe the spatial layout of each semantic part within the object.
The generation process can therefore be written as a mapping
\begin{equation}
    f: C \rightarrow G,
\end{equation}
which we implement through two structured stages:
\begin{equation}
    C \rightarrow (B, V) \rightarrow G
\end{equation}
The first stage predicts identity-aligned spatial layouts, while the second stage synthesizes detailed part geometries conditioned on these layouts.
As illustrated in \cref{fig:overview}, by explicitly anchoring generation to semantic part identities, this formulation establishes a stable correspondence between part identity, spatial layout, and geometry synthesis.

\subsection{Identity-Aware Condition Encoding}
\begin{figure}[tb]
  \centering
  \begin{subfigure}{0.36\linewidth}
    \includegraphics[width=\linewidth]{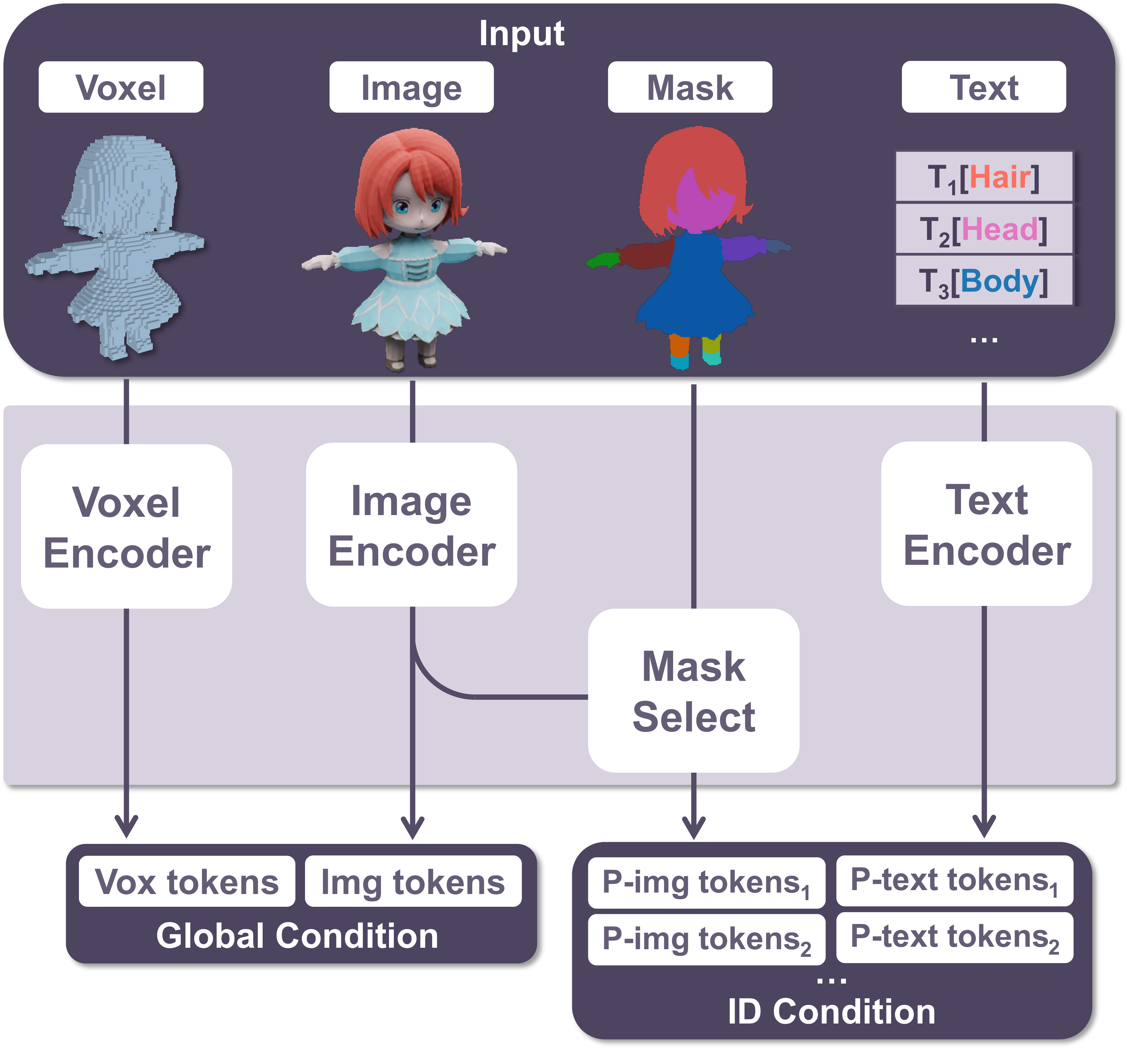}
    \caption{Identity-Aware Condition}
    \label{fig:condition}
  \end{subfigure}
  \hfill
  \begin{subfigure}{0.61\linewidth}
    \includegraphics[width=\linewidth]{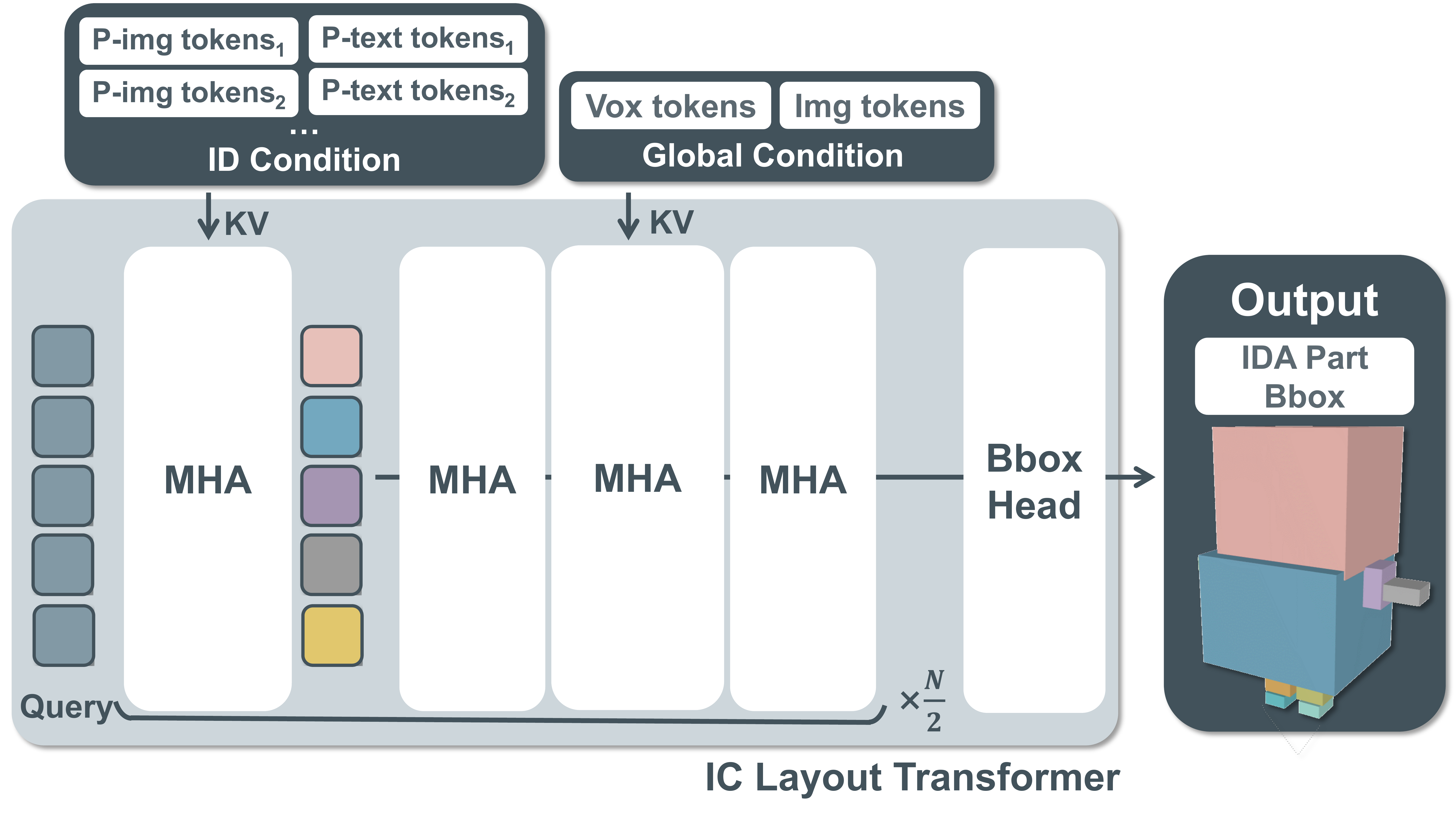}
    \caption{Identity-Conditioned Layout Prediction}
    \label{fig:layout}
  \end{subfigure}
  \caption{
  Identity-aware condition and layout prediction. 
  (a) Multi-modal inputs are encoded into identity conditions and global context tokens.
  (b) An identity-conditioned transformer predicts part bounding boxes using identity-aligned slot queries.
  }
\end{figure}

To enable identity-aligned generation, we construct an identity-aware condition interface that organizes heterogeneous inputs into structured conditions for subsequent spatial reasoning.
As illustrated in \cref{fig:condition}, the input modalities are encoded into per-part identity conditions and shared global context conditions.

\textbf{Local identity conditions.}
For each semantic part, the text description $T_k$ is encoded into a semantic embedding that represents its identity.
When masks are available, mask-guided image features are extracted from the reference image to capture localized appearance cues.
The text embedding and mask-guided image features together form a per-part identity condition $\{\mathcal{LC}_k\}_{k=1}^{K}$, which provides identity-level guidance for each part.

\textbf{Global context conditions.}
In addition to identity-level signals, we construct global conditions that provide object-level context.
The reference image is encoded into image tokens, while the voxel prior $V$ is encoded into voxel tokens representing coarse 3D structure.
These tokens together form the global condition representation $\mathcal{GC}$, which provides holistic context for layout reasoning.

By explicitly separating identity-level conditions from global contextual signals, this interface enables subsequent modules to reason about part identities and spatial structure in a coordinated manner.

\subsection{Identity-Conditioned Layout Prediction}
\label{sec:layout}

Given the identity and global conditions constructed in the previous section, the goal of this stage is to predict spatial layouts aligned with part identities.
To this end, we introduce an identity-conditioned layout transformer that predicts per-part bounding boxes in a parallel and identity-consistent manner (see \cref{fig:layout}).

\textbf{Slot initialization.}
For an object with $K$ semantic parts, we instantiate $K$ parallel prediction queries that serve as layout prediction slots. 
Each slot is responsible for predicting the spatial layout of one semantic part. 
All queries share parameters and are initialized identically, while their behaviours are later differentiated through identity-conditioned attention.

\textbf{Slot–identity alignment.}
To associate each slot with a specific semantic part, we inject the local identity conditions $\{\mathcal{LC}_k\}$ through cross-attention ($\text{MHA}_{id}$).
Each slot attends to its corresponding identity condition, allowing semantic descriptions and mask-guided visual cues to guide layout prediction.
Through this mechanism, the semantic identity of each part directly influences the spatial reasoning process of its corresponding slot.

\textbf{Slot interaction and global grounding.}
The prediction slots further interact with each other through self-attention layers ($\text{MHA}_{s}$), enabling each slot to leverage information from other parts during layout prediction.
In addition, the slots attend to the global condition tokens $\mathcal{GC}$ ($\text{MHA}_{g}$), which encode object-level context derived from the reference image and voxel prior.
This mechanism helps each slot interpret its semantic identity within the overall object structure and align its predicted layout with the global geometry.
Let $\mathbf{q}_k^{(l)}$ denote the $k$-th slot representation at layer $l$. A transformer block updates each slot as
\begin{equation}
\mathbf{q}_k^{(l+1)} =
\mathrm{MHA}_{s}\Big(
\mathrm{MHA}_{g}\big(
\mathrm{MHA}_{s}\big(
\mathrm{MHA}_{id}(\mathbf{q}_k^{(l)}, \mathcal{LC}_k)
\big)
, \mathcal{GC}
\big)
\Big)
\end{equation}
After several transformer blocks performing slot interaction and condition injection, the final slot representations are fed into a bounding box prediction head to produce per-part bounding boxes $\{B_k\}_{k=1}^{K}$.

\subsection{Layout-Conditioned Geometry Synthesis}
\label{sec:geometry}
\begin{figure}[tb]
  \centering
  \includegraphics[width=.97\linewidth]{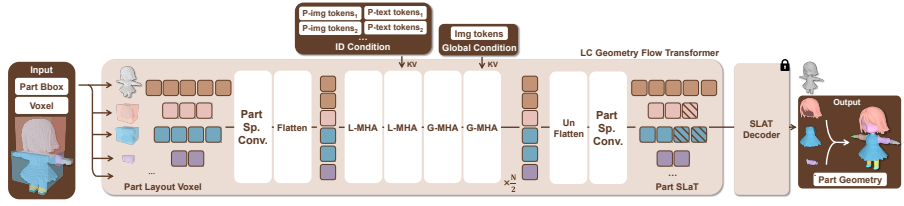}
  \caption{
  Layout-conditioned geometry synthesis. 
  Predicted bounding boxes partition the voxel prior into part layout voxels, which are processed by a geometry flow transformer with local and global attention (L-MHA and G-MHA) to preserve identity consistency across stages.
  Filtered voxels are then decoded to generate the final part geometries.
  }
  \label{fig:geometry_dit}
\end{figure}

Given the identity-aligned layouts predicted in the previous stage, we synthesize detailed part geometries conditioned on these layouts. 

\textbf{Layout-conditioned geometry generation.}
As illustrated in \cref{fig:geometry_dit}, from $K$ predicted bounding boxes $\{B_k\}$, we construct per-part layout voxels by extracting the corresponding spatial regions from the coarse voxel prior.
These layout voxels provide coarse structural anchors for each part and are processed by a geometry flow transformer to predict voxel-wise features and retention probabilities, following the baseline geometry synthesis pipeline.

\textbf{Identity-consistent geometry reasoning.}
Although the layout voxels already encode coarse part structures, we further preserve identity consistency across stages by aligning the geometry network with the identity-aware conditioning interface introduced earlier.
Specifically, the geometry transformer adopts two types of attention blocks.
Local attention blocks (L-MHA) operate independently on each part voxel group and attend to the corresponding identity condition, reinforcing the alignment between voxel features and part semantics.
Global attention blocks (G-MHA) jointly process voxel features from all parts and attend to the global context tokens, enabling spatial grounding and information exchange across parts.

After voxel filtering, the retained voxels are passed to a geometry decoder to generate the final part geometries.
By conditioning geometry generation on identity-aligned layouts, our framework reduces spatial ambiguity and enables stable part-wise geometry synthesis.

\section{Experiment}

\subsection{Dataset}
Learning identity-aligned generation requires consistent supervision for part identities.
However, existing part-level datasets often provide highly descriptive captions that mix geometry, material, and relational attributes, resulting in high semantic entropy and ambiguous identity definitions. 
Such descriptions make it difficult for models to learn stable correspondences between semantic identities and generation slots.

To provide reliable identity supervision, we construct an identity-oriented part-level dataset based on PartVerseXL \cite{ding2025fullpart} with semantic re-annotation.
Instead of free-form captions, we adopt a closed semantic vocabulary for each object category, where each token corresponds to a well-defined part identity. 
To avoid identity ambiguity, we explicitly incorporate directional and cardinal modifiers when necessary (e.g., \textit{left hand}, \textit{right hand}, \textit{both hands}), ensuring that each token uniquely maps to a consistent semantic part across objects.

Initial labels are obtained using a vision–language model constrained by the predefined vocabulary, followed by manual refinement to ensure semantic consistency. 
The resulting dataset contains approximately 8K curated objects with around 70 unified semantic part categories.
This dataset is constructed to enable consistent identity supervision for training and evaluation of identity-aligned generation.
We randomly sample 100 objects for evaluation and the remaining samples for training.

\subsection{Implementation}

We train the layout prediction and geometry synthesis modules in two stages.
The layout stage predicts part bounding boxes using the BBox Transformer introduced in \cref{sec:layout}.
Since no suitable one-to-one layout prior exists, the transformer is trained from scratch with a pretrained voxel encoder providing geometric priors.

For geometry synthesis, we adopt a pretrained flow-based backbone~\cite{yang2025omnipart} and fine-tune it under identity-aware conditioning.
During training, the conditioning bounding boxes are progressively perturbed so that the geometry stage learns to handle imperfect layouts predicted by the layout model.

All models are optimized using AdamW with a learning rate of $1\times10^{-4}$.
The layout transformer is trained with a bounding box classification loss, while the geometry network follows the conditional flow matching objective.
Further details on training objectives and regularization strategies are provided in the supplementary material.

\subsection{Main Comparison}

\begin{figure}
    \centering
    \includegraphics[width=0.97\linewidth]{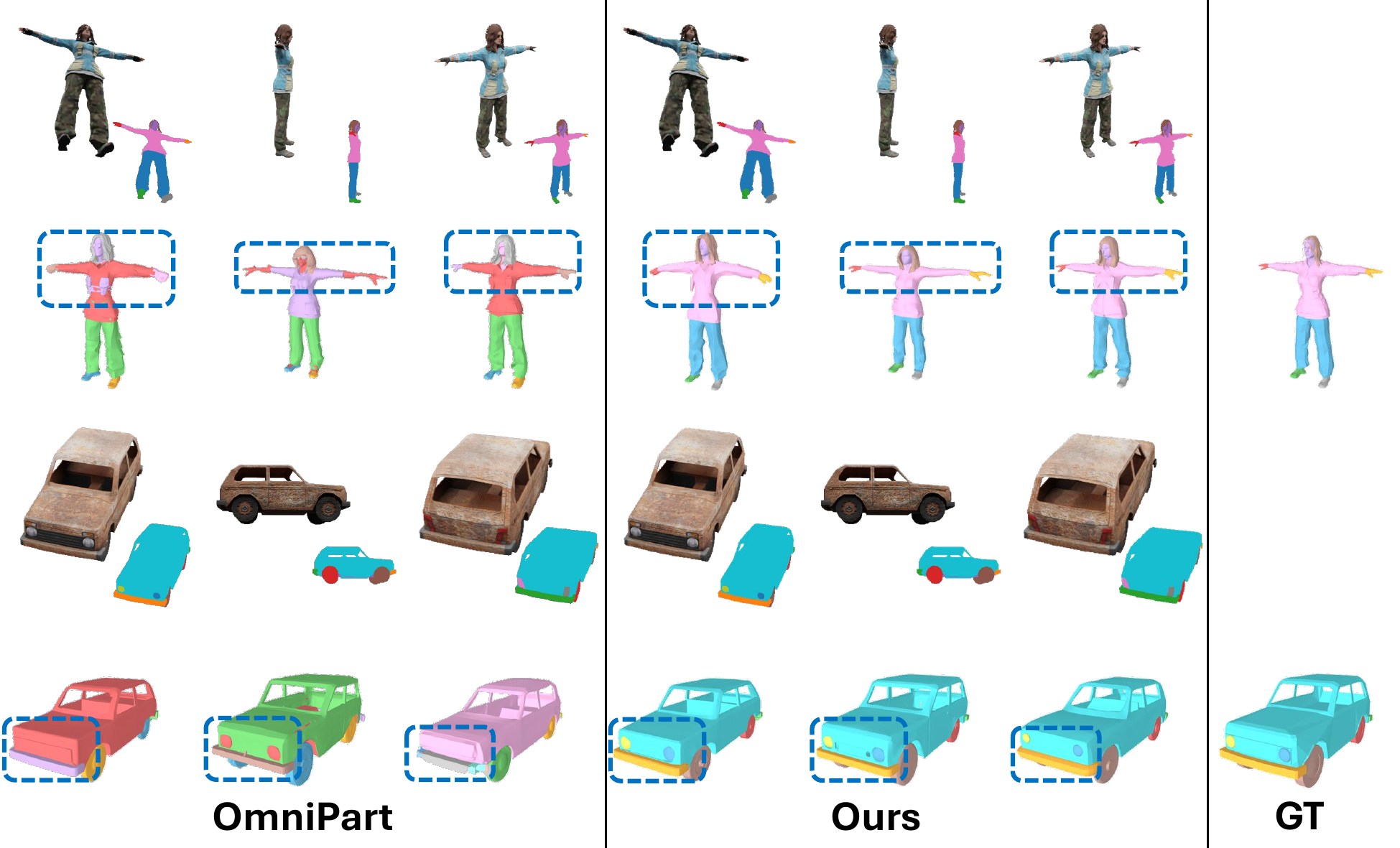}
    \caption{
    Comparison with OmniPart under the same identity condition across viewpoints.
    Colors represent generation slot IDs.
    OmniPart assigns the same semantic part to different slots across viewpoints, leading to inconsistent part identities.
    In contrast, our identity-aligned design preserves stable slot assignments, producing consistent part decompositions.
    }
    \label{fig:omnipart_compare}
\end{figure}

\begin{table}[tb]
  \caption{
  Quantitative comparison of part-aware 3D generation methods.
  We report Chamfer Distance (CD$\downarrow$), F-Score@0.1 ($\uparrow$), Part-IoU ($\downarrow$), and NMI ($\uparrow$).
  Our identity-aligned model achieves the best performance across all metrics.
  }
  \label{tab:quantitative}
  \centering
  \begin{tabular}{@{}lcccc@{}}
    \toprule
    Method & CD $\downarrow$ & F-Score@0.1 $\uparrow$ & Part-IoU $\downarrow$ & NMI $\uparrow$ \\
    \midrule
    PartField~\cite{partfield2025}      & 0.1922 & 0.7158 & 0.0402 & 0.5320 \\
    HoloPart~\cite{yang2025holopart}        & 0.1830 & 0.7278 & 0.0427 & 0.5426 \\
    PartCrafter~\cite{lin2025partcrafter}  & 0.2356 & 0.6268 & 0.0375 & 0.5525 \\
    OmniPart\cite{yang2025omnipart}                    & 0.1621 & 0.8197 & 0.0347 & 0.5632 \\
    Ours                            & \textbf{0.1410} & \textbf{0.8249} & \textbf{0.0330} & \textbf{0.6157} \\
    \bottomrule
  \end{tabular}
\end{table}

\textbf{Baselines.}
We compare our method with four representative part-aware 3D generation approaches that differ in how part identity is modeled during generation.
PartField~\cite{partfield2025} performs post-hoc part segmentation on a holistically generated mesh\cite{Xiang_2025_CVPR}, where part boundaries are inferred purely from geometric cues without explicit identity modeling.
HoloPart~\cite{yang2025holopart} generates part geometry conditioned on segmentation results, but the segmentation itself does not encode stable semantic identities.
PartCrafter~\cite{lin2025partcrafter} directly predicts part-aware shapes from images using compositional latent representations, where part identities emerge implicitly during generation.
OmniPart~\cite{yang2025omnipart} introduces part masks as identity conditions and performs autoregressive layout planning followed by layout-conditioned geometry generation; however, identity signals may be incomplete under occlusion and are not explicitly aligned with dedicated generation slots.
These baselines therefore represent different identity modeling strategies for part-aware 3D generation.

\textbf{Metrics.}
To evaluate part generation from the perspectives of geometric fidelity, structural coherence, and identity alignment, we adopt four complementary metrics.
\textbf{(1)} We measure the bidirectional Chamfer Distance (\textbf{CD}) between the generated full object and the ground-truth mesh.
\textbf{(2)} We compute F-Score with a threshold of 0.1 (\textbf{F-Score@0.1}) to evaluate surface reconstruction quality.
\textbf{(3)} We compute the average pairwise IoU between predicted parts in 3D space. A lower \textbf{Part-IoU} indicates better geometric separation and thus improved structural coherence.
\textbf{(4)} To evaluate identity–slot alignment, we compute Normalized Mutual Information (\textbf{NMI}) between predicted and ground-truth part assignments. 
Points are uniformly sampled on the ground-truth mesh and assigned to their nearest predicted and ground-truth parts. 
NMI measures the consistency between the resulting point partitions and is invariant to label permutations, making it suitable for evaluating identity correspondence.

\begin{figure}
    \centering
    \includegraphics[width=0.9\linewidth]{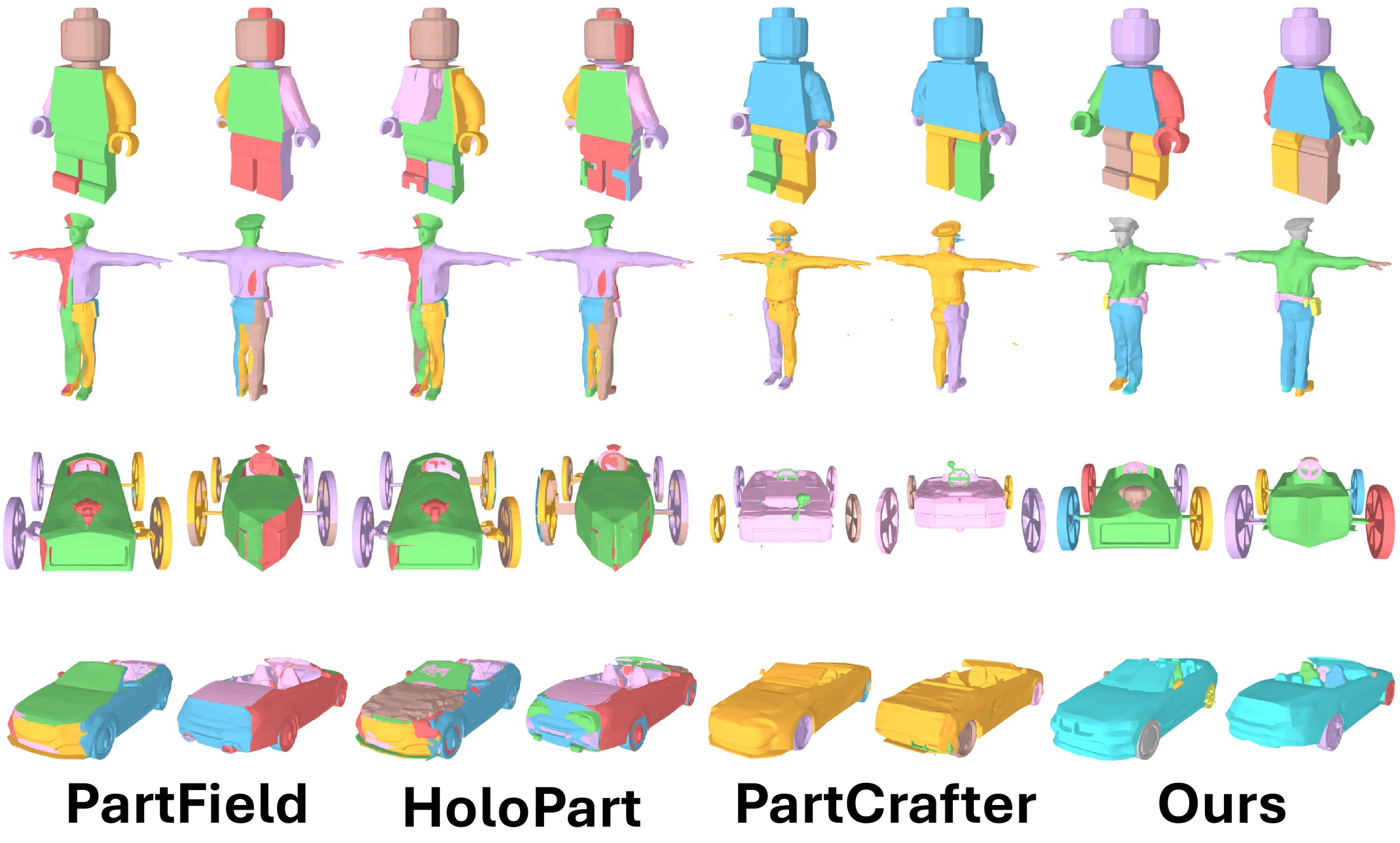}
    \caption{
    Qualitative comparison with representative part-aware generation methods.
    Our identity-aligned model produces semantically coherent and well-separated parts, while prior methods often generate ambiguous or merged components.
    }
    \label{fig:main_qualitative}
\end{figure}

\textbf{Qualitative Results.}
We first compare our method with OmniPart~\cite{yang2025omnipart}, the strongest baseline with explicit identity conditioning, as shown in Fig.~\ref{fig:omnipart_compare}.
We visualize predicted parts using consistent colors based on slot IDs to represent identity–slot assignments.

For the same object and identity conditions, OmniPart produces inconsistent part decompositions across different viewpoints.
In addition, the same semantic part may be assigned to different generation slots, leading to inconsistent color assignments across runs.
This instability arises from two factors.
First, identity cues derived from image masks can become unreliable under occlusion or incomplete observations.
Second, the autoregressive generation process does not explicitly enforce identity–slot alignment, allowing the same semantic part to be assigned to different slots across runs.

In contrast, our method provides stable identity supervision through semantic tokens and explicitly aligns each semantic identity with a dedicated generation slot.
As a result, consistent identity conditions lead to consistent assignments across viewpoints, producing stable and interpretable part decompositions.

We further compare our method with other representative part-aware methods without explicit identity modeling (Fig.~\ref{fig:main_qualitative}).
Across multiple object instances, these methods often produce semantically inconsistent part decompositions due to the lack of explicit identity modeling, leading to ambiguous part boundaries or merged components.
In contrast, our identity-aligned formulation generates semantically coherent parts with clearer boundaries and more stable structures.

\textbf{Quantitative Results.}
Table~\ref{tab:quantitative} presents the quantitative comparison with existing part-aware generation methods.
Our method achieves the best performance across all metrics, demonstrating the effectiveness of identity–slot alignment in stabilizing part-aware generation.

In particular, our model obtains a significantly higher NMI, indicating stronger consistency between predicted and ground-truth part assignments.
This result verifies that explicitly aligning semantic identities with generation slots improves the correspondence between semantic parts and generated components.

The benefits of identity alignment also translate to improved geometric quality.
By reducing ambiguity in part allocation, our formulation produces more coherent structures and better part separation, leading to improved CD, F-Score, and Part-IoU compared with prior approaches.

\subsection{Ablation Study}

\begin{table}[tb]
  \caption{Ablation of training conditions on layout prediction.}
  \label{tab:ablation}
  \centering
  \begin{tabular}{@{}lccc@{}}
    \toprule
    Method & Bbox IoU $\uparrow$ & Voxel IoU $\uparrow$ & Voxel Rec $\uparrow$ \\
    \midrule
    w/o Mask      & 0.4349 & 0.4550 & 0.7538 \\
    w/o Text      & 0.4880 & 0.5111 & 0.8087 \\
    w/o Voxel     & 0.3205 & 0.3393 & 0.5843 \\
    Ours          & \textbf{0.5616} & \textbf{0.5851} & \textbf{0.8818} \\
    \bottomrule
  \end{tabular}
\end{table}

\begin{figure}[tb]
  \centering
  \includegraphics[width=.9\linewidth]{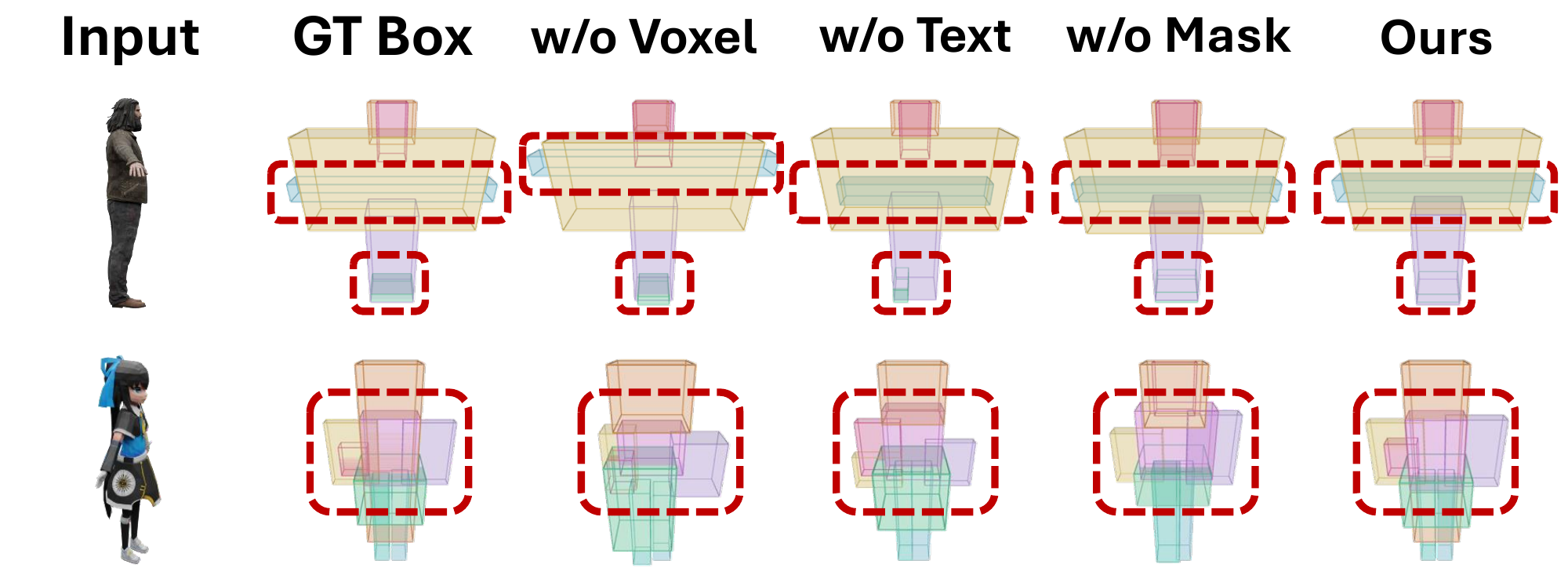}
  \caption{
  Training condition ablation.
  Removing voxel, text, or mask conditions weakens spatial grounding or identity supervision, leading to degraded layout prediction
  }
  \label{fig:condition_ablation}
\end{figure}

\textbf{Training Condition Ablation.}
To analyze how different modalities contribute to identity–slot alignment, we remove individual conditions during training.
Results are reported in Table~\ref{tab:ablation} and Fig.~\ref{fig:condition_ablation}.
BBox IoU measures layout accuracy, while Voxel IoU and Voxel Recall reflect the impact of layout prediction on downstream geometry generation.

Removing voxel features causes the largest performance drop, as the model loses reliable spatial anchors for layout prediction.
Without text identity, the model must rely solely on visual masks, which can become unreliable under occlusion and lead to ambiguous identity assignments.
Removing mask conditions leaves identity supervision entirely to text, which makes it harder to resolve ambiguous boundaries or rare components.

Using all modalities provides complementary identity grounding: text defines semantic identities, masks provide visual localization, and voxels anchor spatial structure, resulting in the most stable identity–slot alignment.

\begin{figure}[tb]
    \centering
    \includegraphics[width=0.9\linewidth]{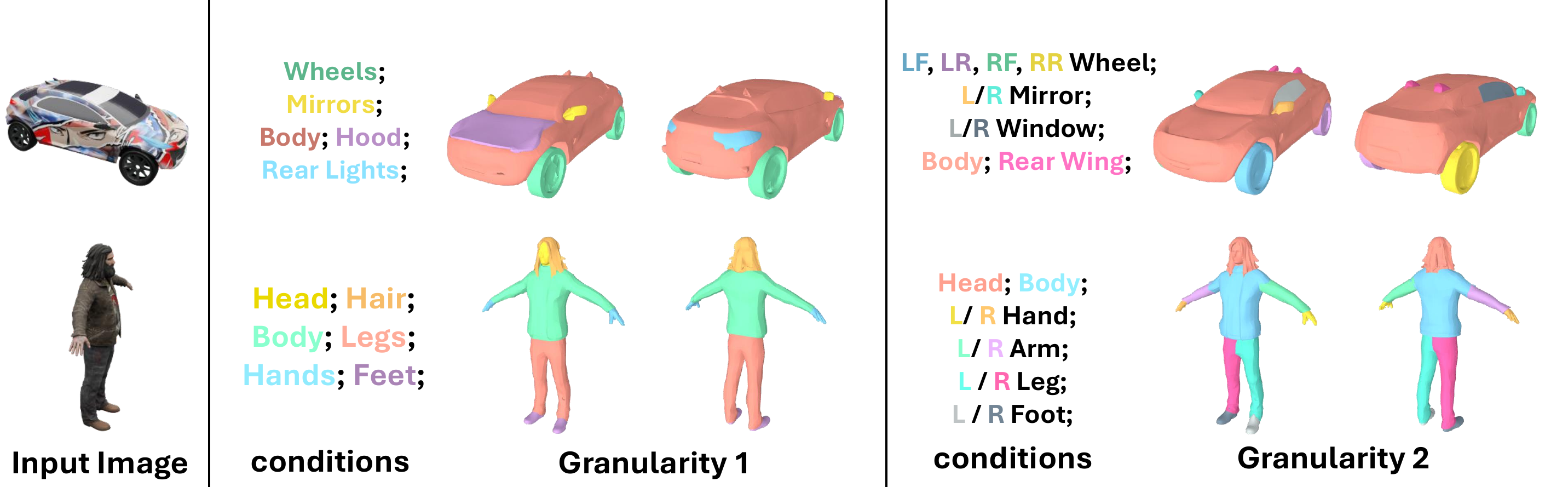}
    \caption{
    Semantic compositional control.
    Different textual identity combinations lead to different part granularities, demonstrating controllable generation enabled by identity–slot alignment.
    }
    \label{fig:granularity_compare}
\end{figure}

\begin{figure}[tb]
  \centering
  \includegraphics[width=.9\linewidth]{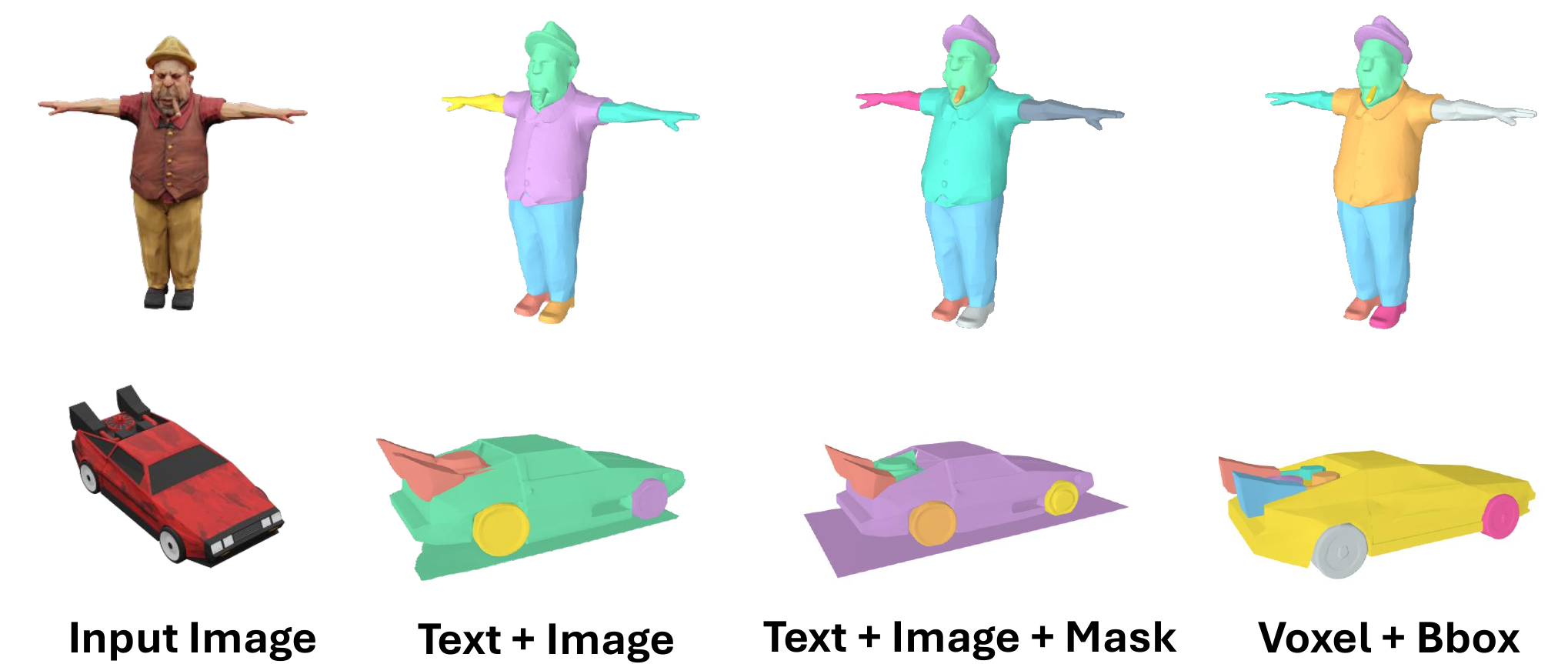}
  \caption{
  Inference condition ablation.
  Our model remains robust under weak conditions (image + text) and supports finer controllability when stronger identity cues such as masks or ground-truth layouts are provided.
  }
  \label{fig:inference_ablation}
\end{figure}

\textbf{Inference Condition Ablation and Granularity Control.}
We further evaluate our framework under different inference-time conditions to analyze the robustness and controllability enabled by identity–slot alignment.
Results are shown in Fig.~\ref{fig:granularity_compare} and Fig.~\ref{fig:inference_ablation}.

With only image and text conditions, the model already produces semantically consistent part decompositions, demonstrating that identity–slot alignment enables reliable generation even without explicit part-level spatial conditions
Moreover, different textual identity combinations naturally lead to different part granularities, showing that the model can adapt the generated structure according to semantic identity specifications.

Introducing mask conditions provides finer-grained identity cues, allowing the model to generate more detailed part compositions, such as small accessories or rare components that are difficult to describe precisely with text alone.

Finally, when ground-truth voxels and part bounding boxes are provided, the model receives the strongest spatial and identity signals, resulting in the highest-quality generation.
Although such conditions are rarely available in practice, they illustrate the upper bound of generation quality under fully specified identity and layout.

Overall, these results demonstrate that explicitly aligning semantic identities with generation slots enables both robust and controllable part-aware generation.

\section{Conclusion}
In this work, we revisit part-aware 3D generation from the perspective of identity–layout entanglement.
We identify identity–slot permutation freedom as a key source of instability, where multiple slot assignments can satisfy the same supervision signal.
To address this issue, we propose identity-aligned one-to-one slot modeling, which explicitly anchors each semantic part to a dedicated generation slot.
Our framework performs identity-conditioned layout prediction followed by layout-conditioned geometry synthesis while preserving identity alignment across stages.
To facilitate training under this paradigm, we construct a part-level dataset with a unified semantic protocol for consistent identity grounding.
Experiments demonstrate that explicit identity–slot alignment significantly improves generation stability, structural consistency, and controllability.
Future work may explore more efficient representations for conveying part identity information.

%
%
\bibliographystyle{splncs04}
\bibliography{main}
\end{document}